\relax
\documentclass[letterpaper]{article} 
\usepackage{aaai21}  
\usepackage{times} 
\usepackage{amssymb}
\newcommand{\qed}{\hfill\blacksquare}
\usepackage[linesnumbered,ruled]{algorithm2e}
\usepackage{helvet} 
\usepackage{courier}  
\usepackage[hyphens]{url} 
\usepackage{graphicx}  
\usepackage{natbib}  
\usepackage{caption} 
\usepackage{amsmath}
\usepackage{amsfonts}
\usepackage{tabularx}
\usepackage{hyperref}
\usepackage{cleveref}
\usepackage{booktabs}
\usepackage[switch]{lineno}
\newcommand{\removelatexerror}{\let\@latex@error\@gobble}

\title{Reinforcement Learning-based N-ary Cross-Sentence Relation Extraction}

\author{
    Chenhan Yuan\textsuperscript{\rm 1}
    Ryan Rossi \textsuperscript{\rm 2}
    Andrew Katz \textsuperscript{\rm 1} 
    Hoda Eldardiry \textsuperscript{\rm 1} 
}
\affiliations{
    \textsuperscript{\rm 1}Virginia Tech\\
    \textsuperscript{\rm 2}Adobe Research \\

    chenhan@vt.edu, ryarossi@gmail.com, akatz4@vt.edu, hdardiry@vt.edu
}

\begin{document}
\maketitle
\begin{abstract}
The models of n-ary cross sentence relation extraction based on distant supervision assume that consecutive sentences mentioning $n$ entities describe the relation of these $n$ entities. However, on one hand, this assumption introduces noisy labeled data and harms the models' performance. On the other hand, some non-consecutive sentences also describe one relation and these sentences cannot be labeled under this assumption. In this paper, we relax this strong assumption by a weaker distant supervision assumption to address the second issue and propose a novel sentence distribution estimator model to address the first problem. This estimator selects correctly labeled sentences to alleviate the effect of noisy data is a two-level agent reinforcement learning model. In addition, a novel universal relation extractor with a hybrid approach of attention mechanism and PCNN is proposed such that it can be deployed in any tasks, including consecutive and non-consecutive sentences. Experiments demonstrate that the proposed model can reduce the impact of noisy data and achieve better performance on general n-ary cross sentence relation extraction task compared to baseline models.
\end{abstract}

\section{Introduction}
As a key step in constructing a knowledge graph, relation extraction is a task to extract the relation between the entities expressed in a sentence. 
Previous work has largely focused on intra-sentence binary relation extraction, where the goal is to extract the relation between an entity pair in the sentence~\cite{hu2019improving,gupta2019neural}. 

However, some relations require more than two entities and may span multiple sentences, which is defined as n-ary cross-sentence relation extraction. As the example shown in Table~\ref{table_exp_oldds}, the relation ``educate'' includes four entities, the person's "name``, "academic degree``, "academic major`` and "school``. In addition, this relation spans in four sentences in the example.
Some prior works have applied a supervised learning approach to tackle this task, but they require large-scale labeled training data~\cite{jia2019document}.
\begin{table}[ht]
\caption{Distant supervision labeled sentences example. Pos. is the sentence position in the text. DS is the relation label result of distant supervision and R is the real label. The fact used here is: \{Alan Turing, PhD, Princeton, computer science\}, which has the ``educate'' relation. ``edu'' denotes that the sentence represents the ``educate'' relation and ``--'' denotes it does not.}
\begin{tabularx}{\linewidth}{ c X c c} \toprule
Pos. & Sentence & DS & R\\ \midrule
3   &   \textbf{Alan Turing} worked on hyper computation in \textbf{Princeton} University.& edu & --  \\
4   &  He obtained his \textbf{PhD} in 1938.  & edu &edu   \\
18  & \textbf{Alan Turing} studied logic and \textbf{computer science} in \textbf{Princeton}.  & -- & edu   \\
20  &  His \textbf{PhD} advisor is \textbf{Alonzo Church} & -- & edu  \\ \bottomrule
\label{table_exp_oldds}
\end{tabularx}
\end{table}

To obtain large-scale annotated data, some work assumes that if the consecutive sentences (a sentence group) contain the entities that have a relation in a knowledge base, these sentences as a whole describe that relation~\cite{quirk2016distant}. 
This assumption is referred to as distant supervision in the n-ary cross-sentence relation extraction task. 
Even though methods based on distant supervision can quickly annotate sentences, they still have two main limitations:
1) they suffer from a noisy labeling problem; 
2) the strong distant supervision assumption does not consider the non-consecutive sentences, which reduces the generalizability of the trained model. As the example shown in Table~\ref{table_exp_oldds}, the sentences at the 18th and 20th positions describe the fact but are not labeled using distant supervision because they are not consecutive. The first sentence is incorrectly labeled and is a noisy labeled data, which describes Alan Turing's work instead of his education.

To address the first limitation, we propose to train a \emph{sentence distribution estimator (SDE)}, which is a two-level agent reinforcement learning model. This provides a well-trained model that can select the high-quality labeled sentence groups and alleviate the impact of noisy data. There are previous works on applying reinforcement learning (RL) to remove binary intra-sentence noisy data and achieve state-of-the-art (SotA) performance~\cite{feng2018reinforcement,yang2019exploiting,qin2018robust}. When applying RL for n-ary cross-sentence relation extraction, a key challenge is that the RL model should not only learn sentence features, but also know the context and relation between each sentence. In this paper, the process of selecting sentences is not only influenced by the feature of the sentence itself, but also by the indicators we defined , which measure the semantic relationship between sentences. Moreover, whether a sentence is selected in a state or not is going to affect the decision of the next state. This state transition property provides the ability to choose the best combination of sentences in each sentence group.

To address the second limitation, 
we relax the strong distant supervision assumption that lies at the heart of prior work by replacing it with a weaker distant supervision assumption. The assumption is that the sentence that has at least one main entity or two supplementary entities is annotated with the relation of these entities. We follow the Wikidata Knowledge Base scheme, where the main entity is the ``value'' of each fact and the supplementary entity is the ``qualifer'' of each fact. This assumption introduces some non-consecutive sentences and we propose a novel universal relation extractor to encode both consecutive and non-consecutive sentence groups. This relation extractor has a self-attention and soft attention mechanism layer, which compares the similarity between the word-level features and the relation query vectors. The relation extractor also encodes each sentence via a Piece-wise Convolution Neural Network (PCNN) layer. The PCNN output is used to learn how the information transforms through sentences via a non-linear transformation layer.

\section{Related Work}
Dependency shortest path has been applied with other pre-processing features for n-ary cross-sentence relation extraction~\cite{li2015improvement,mesquita2013effectiveness}. With the rise of deep learning, some work encoded the dependency shortest path via graph neural networks. Peng et al. applied Graph-LSTM to encode the dependency shortest path and link each path~\cite{peng2017cross}. One dependency shortest path usually requires two Graph-LSTMs. Song et al. proposed the Graph-state LSTM so that only one Graph-LSTM is needed to encode a path~\cite{song2018n}. Some work also implemented Bi-LSTM directly to encode the whole sentence sequences without requiring any preprocessing ~\cite{mandya2018combining}. The LSTM-CNN model they proposed achieved a better performance on the PubMed dataset, but it cannot encode long sequences. Recently, this model has been improved by deploying a multi-head attention layer. The model is also enhanced by incorporating prior knowledge from a pre-trained Knowledge Base~\cite{zhao2020incorporating}.

The large-scale data used in these approaches are automatically labeled via distant supervision. As discussed in previous literature, distant supervision always introduces noisy incorrectly labeled data~\cite{mintz2009distant,takamatsu2012reducing}. In the binary relation extraction task, this problem is addressed by using a weaker distant supervision assumption. This assumption takes all labeled sentences with the same entity pairs into a bag and assumes that only one sentence in this bag is correctly labeled~\cite{ye2019distant,ji2017distant}. Some work also trained an extra selector model, which selected the correctly labeled sentences as the training data of the relation extraction model. Most selectors are reinforcement learning (RL)-based models~\cite{feng2018reinforcement,yang2019exploiting,qin2018robust}. However, these reinforcement learning-based selectors cannot be applied to n-ary cross-sentence relation extraction task, which is more challenging than the binary intra-sentence relation extraction.

To the best of our knowledge, our proposed work is the first to apply reinforcement learning on the n-ary cross-sentence relation extraction task. We propose an RL-based two-level agent selector (sentence distribution estimator) to select the correctly labeled sentence group. We also propose a weaker distant supervision assumption to label both consecutive and non-consecutive sentences. To encode them both, a novel universal relation extractor model is proposed, which is a hybrid approach of attention mechanism and context learning process of sentence features. 
\section{Problem Formulation}
In a relation extraction task, a fact is defined as a collection of $i$ entities and one corresponding relation, where $i\geq 2$. The relations are verb phrases and describe the relationship among these entities. For every $m$ sentences, a relation extraction model should give the relation among entities expressed in these sentences. If $m\geq 1$ and $i\geq 2$, the task is the cross-sentence n-ary relation extraction problem. 

In distant supervision-based method, we decompose the cross-sentence n-ary relation extraction task into two sub-problems: sentence distribution estimation and relation extraction. The sentence distribution estimation is formulated as follows: given a set of sentence groups and relation label pairs as $\{(g_1, r_1), (g_2, r_2), \cdots, (g_n, r_n)\}$, where there are variable numbers of sentences in each sentence group $g$ and $r$ is the noisy relation label produced by distant supervision, the objective is to decide which sentences in each group truly describe the relation. In other words, the model tells which sentence is correctly labeled and should be selected as a training instance. The relation extraction is to classify relation $r$, given a sentence group $g$.
\section{Proposed Model}
\begin{figure*}
\centering
\includegraphics[scale=0.66]{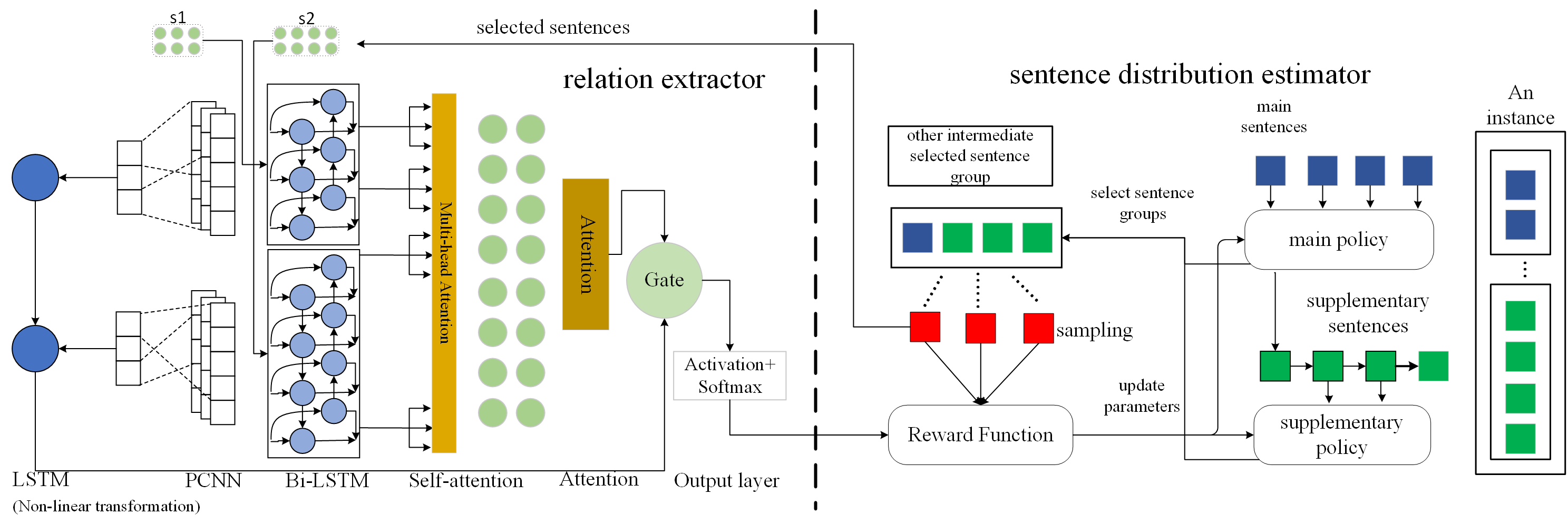}
\caption{The flowchart of the proposed model. On the right side, the sentence distribution estimator consists of main policy and supplementary policy. The relation extraction model is on the left side.}
\label{fig_framework}
\end{figure*}
As shown in Fig \ref{fig_framework}, the proposed model consists of two sub-models. The first is a sentence distribution estimator (SDE), which is used to measure the probability of whether a sentence is correctly labeled by distant supervision. This model outputs a group of sentences that describe a complete fact. The second model is the relation extraction model (RE). This model takes a group of sentences as input and infers the relation contained in these sentences.
\subsection{Sentence Distribution Estimator (SDE)}
The proposed model is a two-level agent reinforcement learning model as in Fig.~\ref{fig_framework}. We assume that for each group of sentences, there is one main sentence and some supplementary sentences. The main sentence that has main entities may give the main information about the relation. The supplementary sentences supplement this information and are selected given the main sentence.
\subsubsection{Main Sentence-Level Policy}
\textbf{State} The vector representation of main sentence $i$ is state $s_i$, and is generated by the PCNN layer of the RE model.\\
\textbf{Action} The action set for this level is $a_i\in\{0,1\}$, where $1$ indicates that the program selects the sentence $i$ as the correctly labeled sentence. Note that this is a one-state RL and the reward is calculated once $a_i$ is decided.\\
\textbf{Policy} The policy $\pi_\theta$ represents the probability of selecting the input sentence given the  encoding information $s_i$. 
\begin{equation}
\begin{aligned}
\pi_\theta(a_i, s_i)&=P(a_i|s_i)\\
&=\sigma(W^\top s_i + b)
\end{aligned}
\end{equation}where $\sigma$ is the sigmoid function and $W^\top \in \mathbb{R}^{d_s\times 1}$ is the weighting matrix and $d_s$ is the dimension of the state vector.\\
\textbf{Reward} The reward is the classification accuracy of the relation extraction model, given selected input sentences.\\
\subsubsection{Supplementary Sentence-Level Policy}
\label{sec:sup_policy}
\textbf{State} The state $m_j$ for this level comprises three indicators and the encoding information of sentence $c_j$. The first indicator $e^{-d}$ measures the distance between the current supplementary sentence and the main sentence, where $d$ is the position distance. Position distance is defined as the number of sentences between the two sentences in the text. The second indicator $|\{e|e\in sent_j \land e\in E\}|$ gives the variety of entities of the current sentence $sent_j$, where $E$ is the set of entities of the fact. The third indicator $\frac{c_j\cdot s_i}{||c_j||_2\cdot||s_i||_2}$ measures the cosine similarity between the current sentence and the main sentence $i$. Along with the encoding information, we assume that these indicators can fully address the context information when selecting supplementary sentences.

This is a multi-state RL. The transition function between each $m_j$ should be defined to calculate the reward once the end state is reached. The transition function is defined as follows. We first sort the supplementary sentences according to their corresponding second indicators, then let the agent decide if selecting the first sentence. The next state is the first sentence of the sorted remaining supplementary sentences according to $|\{e|e\in sent_j \land e\in E/E_{prev}\}|$, where $E_{prev}$ is the entities in the previous selected sentences.\\
\textbf{Action} The action set for this level is $b_j\in\{0,1\}$, where $1$ indicates that the program selects the current sentence as the correctly labeled sentence, where the label is the relation indicated by the sentence.\\
\textbf{Policy} Eq.\ref{sup_policy} shows the policy $\pi_\gamma$ considers sentence-level indicators and sentence encoding information simultaneously.
\begin{equation}
\begin{aligned}
        \pi_\gamma(b_j, m_j)&=P(b_j|m_j)\\
        &=\sigma\left(\alpha(W_k^\top k_j + b_k) + \beta(W_s^\top s_j + b_c)\right)\\
 \end{aligned}
 \label{sup_policy}
\end{equation} where $W_k\in \mathbb{R}^{3\times 1}$ and $W_s\in \mathbb{R}^{d_s\times 1}$ are weighting matrices. $k_j$ is the vector of three real-number indicators and $d_s$ is the dimension of the encoding vector of the sentence. $\alpha$ and $\beta$ are also learnable parameters.\\
\textbf{Reward} Note that the rewards, a.k.a the accuracy of the results from the RE model, can only be calculated when all necessary sentences are given. Therefore, there is no intermediate reward that can be used directly for updating gradients in this level policy. Similar to playing Go, we apply the Monte Carlo search algorithm to simulate possible future results and use the average of these results as the intermediate reward~\cite{silver2016mastering}. More formally, given the current state and previous states $m_{1:j}$, the Monte Carlo search algorithm with $\pi_\gamma$ as roll-out policy is applied to sample the future possible state transitions $m'_{j+1:M}$, where $M$ is the end state. The mathematical definition is in Eq.\ref{MC}
\begin{equation}
\begin{aligned}
&Monte_{\pi_\gamma}(m_{1:j};m'_{j+1:M};N)=\\
&\{m_{1:j}m_{j+1:M}^{'(1)},m_{1:j}m_{j+1:M}^{'(2)},\dots, m_{1:j}m_{j+1:M}^{'(n)}\}
\end{aligned}
\label{MC}
\end{equation} where $N$ is sample times. Based on this, the intermediate reward can be calculated via Eq.\ref{inter_reward}
\begin{equation}
\begin{aligned}
&R(i,j)=\\
&\left\{
\begin{array}{ll}
\frac{1}{N}\sum_{n=1}^{N} e^{-Ce(RE(s_i,m_{1:M}^{(n)}))}, m_{1:M}^{(n)}\in Monte_{\pi_\gamma}     & j<M\\
e^{-Ce(RE(s_i,m_{1:j}))}     &j=M\\
\end{array} \right.
\end{aligned}
\label{inter_reward}
\end{equation} where $Ce$ denotes the cross entropy loss and $RE$ is the relation extraction model. Note that an exponential function is applied to make sure that the sentences group that has lower cross entropy loss has a greater reward value.
\subsubsection{Policy Gradient}
After agents decide a set of actions based on their policies, the relation extraction model gives the rewards. The objective of an RL algorithm is to maximize the overall rewards by updating policy parameters following a policy gradient strategy~\cite{sutton2000policy}. The gradients of our two agents can be computed using Eq.\ref{gradient}.
\begin{equation}
\begin{aligned}
        \bigtriangledown \overline{R}_{\theta}&= \sum_{\tau}\sum_{\iota}R(\iota)\pi_\gamma(\tau,\iota)\pi_\theta(\tau)\bigtriangledown log\pi_\theta(\tau)\\
        &= \mathbb{E}_{\tau \sim \pi_\theta(\tau)}[\sum_{\iota}R(\iota)\pi_\gamma(\tau,\iota)\bigtriangledown log \pi_\theta(\tau)]\\
        &\approx \frac{1}{N} \sum_{i=1}^{N}\sum_{j=1}^{M}R(i,j)\pi_\gamma(b_{i,j}, m_{i,j})\bigtriangledown log\pi_\theta(a_i, s_i)\\
        \bigtriangledown \overline{R}_{\gamma}&= \sum_{\tau}\pi_\theta(\tau)\bigtriangledown \sum_{\iota}R(\iota)\pi_\gamma(\tau,\iota) log\pi_\gamma(\tau,\iota)\\
        &= \sum_{\tau} \pi_\theta(\tau) \mathbb{E}_{\iota \sim \pi_\gamma(\iota)}[R(\iota) \bigtriangledown log\pi_\gamma(\iota)]\\
        &\approx \sum_{i=1}^{N}\pi_\theta(a_i, s_i)\frac{1}{M}\sum_{j=1}^{M}R(i,j)\bigtriangledown log\pi_\gamma(b_{i,j}, m_{i,j})\\
 \end{aligned}
 \label{gradient}
\end{equation} where $\bigtriangledown \overline{R}_{\theta}$, $\bigtriangledown \overline{R}_{\gamma}$ denote the derivation of the reward $R$ w.r.t. parameters of main sentence policy $\pi_\theta$ and parameters of supplementary sentence policy $\pi_\gamma$, respectively. Then the parameters of $\pi_\theta$ and $\pi_\gamma$ can be updated via Eq.\ref{update}
\begin{equation}
    \begin{aligned}
    &\theta \leftarrow \theta + \alpha_\theta \bigtriangledown \overline{R}_{\theta}\\
    &\gamma \leftarrow \gamma + \alpha_\gamma \bigtriangledown \overline{R}_{\gamma}
    \end{aligned}
\label{update}
\end{equation}where $\alpha_\theta$ and $\alpha_\gamma$ are the learning rates.\\

\subsection{Relation Extraction Model (RE)}
As shown in Fig~\ref{fig_framework}, the relation extraction (RE) model receives a group of sentences from SDE and then encodes them using a Bidirectional-LSTM layer. We implement attention and self-attention mechanisms to incorporate these sentence encodings to classify the relation. Conventional cross-sentence relation extraction models encode consecutive sentences so they do not consider the connection and context between sentences, which should be learned when encoding non-consecutive sentences. Therefore, attention mechanism is enriched with the output from the non-linear transformation (LSTM) layer by a gate layer in the proposed model. This layer learns how the information transforms in each sentence, which is exactly the context information. The input of this transformation layer is the Piecewise Convolutional Neural Network (PCNN), which encodes each sentence as a feature vector. By applying the hybrid model of attention mechanism and non-linear transformation, the proposed model is universal for cross-sentence n-ary relation extraction task in many scenarios, including both non-consecutive and consecutive sentences.   
\subsubsection{Sentence Encoding}
Bidirectional-LSTM is applied as the sentence encoding layer~\cite{graves2005framewise}. The input of Bi-LSTM is a sequence of the concatenation of the word embedding vector and position encoding vector. We pre-train the word embedding with $d_w$ dimension by Word2Vec~\cite{mikolov2013distributed}. The position encoding is implemented as follows. Supposing the ordered entity list of one sentence is $e_1e_2\dots e_n$, we calculate the position distances between words and $e_1,e_n$. Then these position distances are projected to a dense vector space, which has dimension $d_p$. Although we can select position distances of all entities, more position encoding features will decrease the classification accuracy~\cite{mandya2018combining}. The input dimension for Bi-LSTM of one sentence is $\mathbb{R}^{n_w\times(d_w+d_p+d_p)}$, where $n_w$ is the number of words in each sentence. Then the output dimension from the Bi-LSTM of one sentence is $\mathbb{R}^{n_w\times d_b}$, where $d_b$ is the hidden dimension of the Bi-LSTM.
\subsubsection{PCNN and Non-linear Transformation Layer}
The vector representation of each sentence is used in the non-linear transformation layer. We apply PCNN to take the output of Bi-LSTM layer and get the vector representation~\cite{zeng2015distant}. PCNN first uses $n_f$ filters, each of kernel size $\mathbb{R}^{n_s\times d_b}$, to extract features, where $n_s$ is the window size. The output of each filter $f_i$ is then divided into three segments $\{f_{i1},f_{i2},f_{i3}\}$ according to entities $e_1,e_n$ positions, and does max-pooling in segments. Eq.\ref{pcnn} formally defines the piece-wise max-pooling layer:
\begin{equation}
\begin{aligned}
p_{ij} &= max(f_{ij}), 1\leq i \leq n_f, j=3\\
p_i &= p_{i1}\oplus p_{i2}\oplus p_{i3}\\
 \end{aligned}
 \label{pcnn}
\end{equation} where $\oplus$ denotes concatenation and $p_i\in \mathbb{R}^{1\times3}$ is the piece-wise max pooling result of $i$-th filter. Then the output dimension of PCNN for one sentence is $\mathbb{R}^{3n_f}$.

The non-linear transformation layer is implemented with LSTM Cell~\cite{hochreiter1997long}. The sentence feature vector $s_i$ coming from the PCNN layer is the input at each LSTM cell state. The hidden vector of the last state is the output of the non-linear transformation layer. The mathematical definition is shown in Eq.\ref{non_linear}
\begin{equation}
\begin{aligned}
h_i, c_i &= LSTM(s_i, h_{i-1}, c_{i-1}),  1\leq i \leq n_{se}\\
h_0, c_0 &\sim \mathcal{N}(0, 1)\\
q &= h_{n_{se}} \\
 \end{aligned}
 \label{non_linear}
\end{equation} where $\mathcal{N}(0, 1)$ denotes standard normal distribution and $n_{se}$ is the number of sentences in the sentence group. $q\in \mathbb{R}^{1\times d_h}$ is the output of non-linear transformation layer, where $d_h$ is the hidden dimension of LSTM cell.
\subsubsection{Attention and Self-attention Mechanism}
Previous work reports that multi-head self-attention improves sentence-level relation extraction performance because of its ability to model long sequences~\cite{zhao2020incorporating, vaswani2017attention}. This mechanism is applied via Eq.\ref{attention_self}:
\begin{equation}
\begin{aligned}
M_i &= softmax\left(\frac{QW_i^Q(KW_i^K)^\top}{\sqrt{d}}\right)VW_i^V\\
M &= M_1\oplus  M_2\oplus  M_3\oplus \cdots M_{n_{he}}\\
U &= MW^O\\ 
 \end{aligned}
 \label{attention_self}
\end{equation} where $W_i^Q\in\mathbb{R}^{n_{se}\times \frac{d_s}{n_{he}}}$, $W_i^K\in\mathbb{R}^{n_{se}\times \frac{d_s}{n_{he}}}$, $W_i^V\in\mathbb{R}^{n_{se}\times \frac{d_s}{n_{he}}}$, $W_i^O\in\mathbb{R}^{d_s\times d_s}$ are learnable parameters and $Q\in\mathbb{R}^{n_{se}\times d_s}$,$K\in\mathbb{R}^{n_{se}\times d_s}$,$V\in\mathbb{R}^{n_{se}\times d_s}$ are the query, key and value vectors projected from the input vectors. $n_{he}$ and $d_s$ are the number of heads and the number of hidden units, respectively.

Another soft attention layer is applied to attend to the input $U$ that contributes the most on the classification of the relation. As shown in Eq.\ref{attention}, this layer compares the relation vectors with output vectors from multi-head self-attention. 
\begin{equation}
\begin{aligned}
p_k &= \sum_{j=0}^{m}\epsilon_{k,j}u_j\\ 
\epsilon_{k,j}&= \frac{e^{c_{k,j}}}{\sum_{i=0}^{m}e^{c_{k,i}}}\\
c_{k,j}&=r_ku_j^T\\
 \end{aligned}
 \label{attention}
\end{equation} where $r_k\in\mathbb{R}^{d}$ is the learnable vector of k-th relation and $p_k\in\mathbb{R}^{d}$ is the attention result for relation $r_k$. 
\subsubsection{Gate Layer and Output Layer}
As shown in Eq.\ref{gate}, an element-wise gate layer is applied to incorporate the outputs from attention layer and non-linear transformation layer.
\begin{equation}
\begin{aligned}
\alpha &= \sigma(W_a^\top S_a+b_a)\\
\tilde{S_n} &= tanh(W_n^\top S_n+b_n)\\
S &= \alpha S_a+(1-\alpha) \tilde{S_n}\\
 \end{aligned}
 \label{gate}
\end{equation} where $S_a \in\mathbb{R}^{d_s}$ is the attention result and $W_a \in\mathbb{R}^{d_s}$ is the weighting matrix. $S_n \in\mathbb{R}^{d_s}$ is the LSTM's result and $W_n \in\mathbb{R}^{d_s}$ is the weighting matrix.
\subsection{Model Training}
The RE and SDE models are iteratively trained. This is based on the following proposition: ``The proposed model can accurately classify correctly labeled data and remove the incorrectly labeled data via this training process.'' This can be formally stated in Proposition 1 below.\\
\textbf{Proposition 1.} After iteratively training SDE and RE, the probability $P_{\theta,\gamma}(1|x_p)$ of selecting sampled $x_p$ as the true labeled data is defined as follows:
\begin{equation}
    P_{\theta,\gamma}(1|x_p) \gg P_{\theta,\gamma}(1|x_n)
\end{equation} 
where $x_p$ represents the correctly labeled training data and $x_n$ represents the incorrectly labeled data. We provide a proof sketch of this proposition in the appendix. The core proof idea is that if the number of samples of $x_p$ is far greater than that of $x_n$, the RE model will converge to the $x_p$ distribution. In other words, the data assigned to the high reward is generally also correctly labeled data. Then SDE will assign higher probability of being selected to high reward data in each iteration. In the end, the data assigned high probability is the correctly labeled data (Proposition 1).
\begin{table*}[ht]
\caption{An example of using the weaker distant supervision to label WikiText, given the fact from Wikidata Knowledge Base}
\centering
\begin{tabular}{l|c}
\hline
Fact & \begin{tabular}[c]{@{}c@{}}\{relation: educated at; main entities: Marie Curie, University of Paris;\\ supplementary entities: physics(major), Doctor of Science(degree)\}\end{tabular}\\ \hline
main sentence & In June 1903, \textbf{Marie Curie} was awarded her doctorate from the \textbf{University of Paris}.\\
main sentence  &  \textbf{Marie Curie} was the first woman to become a professor at the \textbf{University of Paris}.\\
supplementary sentence & \begin{tabular}[c]{@{}c@{}} In 1893, \textbf{Marie Curie} was awarded a degree in \textbf{physics} and \\began work in an industrial laboratory of Professor Gabriel Lippmann.\end{tabular}\\ \hline
\end{tabular}
\label{table_exp_ds}
\end{table*}
\section{Experiments}
\begin{table*}[ht]
\centering
\caption{Average test accuracy in five-fold validation on PubMed dataset. Ternary denotes drug-gene-mutation interactions and Binary denotes binary drug-mutation interactions. ``-'' denotes that the value is not provided herein}
\begin{tabular}{lccccccl}
\toprule
Model & \multicolumn{4}{c}{Binary class}  & \multicolumn{2}{c}{Multi-class}             &  \\ 
\cmidrule{2-7}
                       & \multicolumn{2}{c}{Ternary}                 & \multicolumn{2}{c}{Binary}                  & Ternary              & Binary               &  \\ 
\cmidrule{2-7}
                       & Single               & Cross                & Single               & Cross                & Cross                & Cross                &  \\ 
\midrule
Graph LSTM-EMBED\cite{peng2017cross}       & 76.5                 & 80.6                 & 74.3                 & 76.5                 & ---                  & ---                  &  \\
Graph LSTM-FULL\cite{peng2017cross}        & 77.9                 & 80.7                 & 75.6                 & 76.7                 & ---                  & ---                  &  \\
Graph LSTM MULTITASK\cite{peng2017cross}   & ---                  & 82                   & ---                  & 78.5                 & ---                  & ---                  &  \\
LSTM-CNN\cite{mandya2018combining}               & 79.6                 & 82.9                 & 85.8                 & 88.5                 & ---                  & ---                  &  \\
GCN (K=0)\cite{zhang2018graph}              & 85.6                 & 85.8                 & 82.8                 & 82.7                 & 75.6                 & 72.3                 &  \\
GS GLSTM\cite{song2018n}               & 80.3                 & 83.2                 & 83.5                 & 83.6                 & 71.7                 & 71.7                 &  \\
AGGCN\cite{zhang2019attention}                   & 87.1                 & 87                   & 85.2                 & 85.6                 & 79.7                 & 77.4                 &  \\
Multihead attention\cite{zhao2020incorporating}    & 81.5                 & 87.1                 & 87.4                 & \textbf{89.3}                 & 84.9                 & 80.1                 &  \\ 
\midrule
RE model(ours)         & 88.0$\pm$0.3                & 88.3$\pm$0.2                & 89.1$\pm$0.2                     & 86.5$\pm$0.4                 & 85.1$\pm$0.3                 &      80.4$\pm$0.2                &  \\
RE with SDE(ours)             & \textbf{88.6}$\pm$0.1  &  \textbf{89.2}$\pm$0.1  & \textbf{90.1}$\pm$0.2  & 88.7$\pm$0.3   & \textbf{86.7}$\pm$0.2                  &  \textbf{81.3}$\pm$0.2     &  \\ 
\bottomrule
                       & \multicolumn{1}{l}{} &  \multicolumn{1}{l}{} &  \\

\end{tabular}
\label{table_pub}
\end{table*}

\subsection{Datasets}
\subsubsection{PubMed} The PubMed dataset is created by automatically labeling biomedical literature with Gene Drug Knowledge Database. The labeling process follows this rule: a candidate is retained only if no other co-occurrence of the same entities in an overlapping text span with a smaller number of consecutive sentences~\cite{peng2017cross}. In this dataset, there are 6,987 ternary drug-gene-mutation relation instances and 6,087 binary drug-mutation relation instances. There are 5 categories of relations:``resistance or nonresponse'', ``sensitivity'', ``response'', ``resistance'' and ``none''. Following previous work~\cite{peng2017cross}, we binarize the multi-class relations by replacing the first four relations as ``yes''. We report the experimental results on the binary relation extraction and on the multi-class relation extraction.
\subsubsection{WikiText} A complete fact not only appears in consecutive sentences but also in non-consecutive sentences. The strong distant supervision hypothesis used in PubMed only consider consecutive sentences. To consider these two situations at the same time and test whether the proposed model can reduce the impact of noise data in both situations, we also create a new dataset using a weaker distant supervision assumption.\footnote{en.wikipedia.org/wiki/Wikipedia:Contents/People\_and\_self}
We first collect Wikipedia webpages under the ``People'' category and remove all non-text symbols~\cite{vrandevcic2014wikidata}. Then Wikidata is used as a Knowledge Base to automatically label the relations for these webpages. In Wikidata, each fact consists of two values (main entities), $n$ qualifiers (supplementary entities) with $n$ roles where $n\geq0$, and one property (relation). The labeling process follows this rule: if the sentence has at least one main entity or two supplementary entities that participate in one specific fact, this sentence possibly indicates the relation of that fact. Specifically, as the example shown in Table~\ref{table_exp_ds}, if the sentence has two main entities, this sentence is labeled as the main sentence of that relation. Others are labeled as supplementary sentences of that relation. Note that using this labeling process, some sentences may be labeled more than one relation, which makes the task more challenging. Compared to distant supervision used in the PubMed dataset, this labeling process is a weaker distant supervision assumption and does not restrict the consecutiveness.

Statistically, there are 2,133 facts, 4,194 main sentences and 13,440 supplementary sentences in the WikiText dataset. The number of different relations is 55, while the number of different roles is 90. We select 20\% main sentences and 20\% supplementary sentences individually as the test dataset. In this randomized selection, we also make sure that the instance that has sentences in the test dataset also has sentences in the training dataset. This selection process is applied for 5 times and we report the average accuracy and standard derivation on this dataset.

\subsection{Experimental Settings}
Following previous work~\cite{zhao2020incorporating}, the hyper-parameters are decided based on preliminary experiments on a small development set. For the RE model, we set the embedding size of words and positions to 200 and 25, respectively. The hidden dimension of Bi-LSTM is 252 and the number of filters of PCNN is 132. The window size of PCNN is 5. For SDE model, the sample times for Monte Carlo search is 5 and the sample times for supplementary policy is 3. The learning rate is set to 0.001 and the batch size is set to 128. In WikiText dataset, for every 20 times RE training, we train SDE 8 times. To avoid overfitting, the l2 norm is adopted with ratio 0.1 and the dropout is applied on the Bi-LSTM layer and the output layer with probability 0.5. 

The settings of baseline models are the same as their papers. Note that there are no main sentences and supplementary sentences in PubMed dataset. Therefore, we only implement main sentence-level agent on this dataset, which decides the correct labeling instances. we use Pytorch (v1.6.0) and Numpy to implement the proposed model and generate random numbers and the random seed is set to 10~\cite{paszke2019pytorch,van2011numpy} Our platform is a Tesla P40 GPU with 12G memory.
\subsection{Models}
On PubMed dataset, the proposed model is evaluated against following baseline models: (a) Graph LSTM-based models, including Graph LSTM-EMBED/FULL/ multitask~\cite{peng2017cross}; (b) Graph state LSTM model(GS LSTM)~\cite{song2018n}; (c) LSTM-CNN model, which encodes sentences using LSTM first then extracts features using CNN~\cite{mandya2018combining}; (d) Graph Convolutional Networks (GCN) and Attention Guided GCN (AGGCN)~\cite{zhang2018graph,zhang2019attention}; (e) Multi-head attention-based model model~\cite{zhao2020incorporating}. Besides baselines, the RE model is also tested individually as a variant of the proposed model.

On WikiText dataset, since previous models do not address how to encode non-consecutive sentences, these models cannot be directly applied on this dataset. Therefore, we select two SotA models, multi-head attention and LSTM-CNN, as the baseline models and implement them on WikiText dataset. We also report the performance of variants of the proposed model, which are: (a) the RE model only; (b) the proposed model with randomly supplementary sentence selection; (c) the proposed model without three indicators.
\subsection{Results}
\subsubsection{Evaluation on PubMed}We report average test accuracy in five-fold validation on the PubMed dataset. As shown in Table~\ref{table_pub}, the performance of the proposed model is better than previous SotA baselines on most tasks. Specifically, the test accuracy of RE model on all ternary relation tasks is higher than baselines, which shows that the RE model is capable of processing multi-entity relation extraction. After training the SDE model and the RE model iteratively, the impact of noise data on the training process of the RE model is greatly reduced, so that the accuracy of the proposed model is higher than that of the RE model on all tasks. Meanwhile, we notice that the accuracy of most baselines on multi-class tasks is much lower than that on binary-class tasks, e.g., the accuracy of AGGCN is reduced by about 10\%. However, our model still maintains a high accuracy even on multi-class tasks, which is 1.8\% higher than SotA.

In the binary entity relation extraction tasks, the performance of our model drops a little. One possible reason is that we apply PCNN to extract the feature of each sentence. In the binary relation data, there are many sentences with only one entity, which does not meet the conditions of PCNN. In the experiment, the second anchor of PCNN on this kind of sentence is set at the beginning of the sentence by default.
\begin{table}[ht]
\centering
\caption{The average test accuracy and standard deviation on WikiText dataset.}
\begin{tabular}{lc}
\toprule
\multicolumn{1}{l}{Model} & Accuracy(\%) \\ 
\midrule
LSTM-CNN  & 37.9 $\pm$ 2.5\\
Multihead Attention      & 52.2 $\pm$ 2.6        \\
RE                        & 59.3  $\pm$ 1.7       \\
RE with SDE(random)        & 64.7  $\pm$ 1.3       \\
RE with SDE(no indicators)  & 65.1 $\pm$ 0.8\\
RE with SDE               & \textbf{66.4} $\pm$ 0.9        \\ \bottomrule
\end{tabular}
\label{table_wiki}
\end{table}

\subsubsection{Evaluation on WikiText} Since the baseline model is designed for consecutive sentences, the order of input sentences is set so that the main sentence is the first, followed by all the supplementary sentences in order. This order is also used in our proposed model without the SDE model. 

As shown in Table~\ref{table_wiki}, the test accuracy of the RE model is 7.1\% higher than the best performance of baselines. This shows that the RE model is more capable of encoding non-consecutive sentences and predicting the relations than previous models. Considering both the results on the WikiText and PubMed dataset, the proposed RE model is a universal model that fits for both non-consecutive and consecutive cross-sentence n-ary relation extraction tasks. Note that the number of relations (classes) is 55, so the 66.4\% test accuracy of the proposed model is a fairly great result, which is significantly better than the RE model. This indicates that with the help of SDE agents, the RE model is more possible to learn the real relation distribution.
\subsubsection{Evaluation on SDE model}
\begin{figure}[ht]
\centering
\includegraphics[scale=0.33]{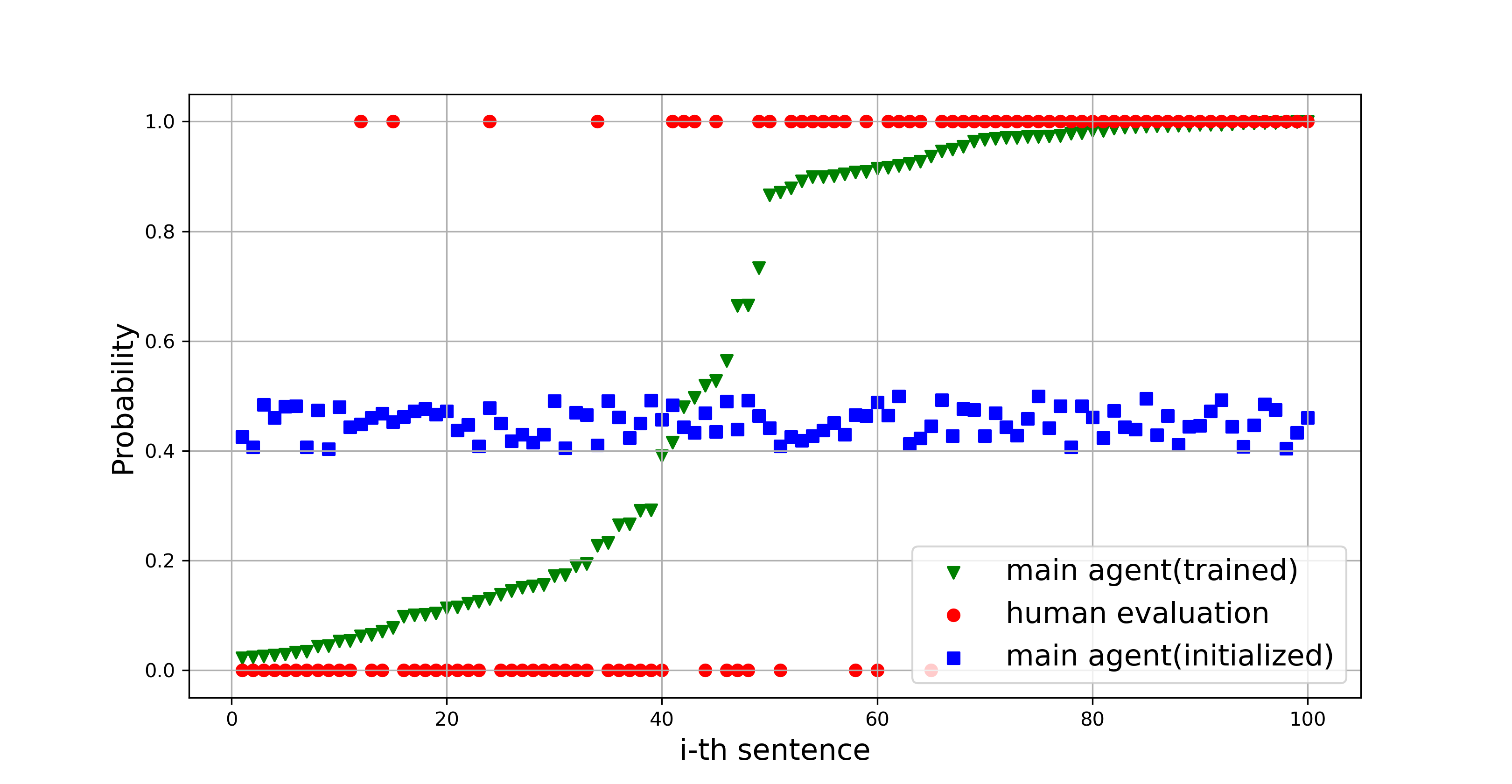}
\caption{The probability distribution of sentences}
\label{fig_mainrl}
\end{figure}
We first randomly select 100 main sentences from the test set and ask a graduate student to check whether the relation labeled with distant supervision is correct for these 100 sentences. The correctly labeled sentences are marked with ``1'', while others are marker with ``0''. Given by the main sentence-level agent, the probability that the sentences are correctly labeled is also reported. As shown in Fig.\ref{fig_mainrl}, the probability distribution of the sentences given by the agent has a strong positive correlation with the results of manual inspection. Specifically, most low-probability sentences are marked with ``0''  by human evaluation, which indicates that these sentences are incorrectly labeled by distant supervision, while most high-probability sentences are correctly labeled. This demonstrates that the well-trained main sentence-level agent can distinguish incorrectly labeled data from correctly labeled data.

\begin{figure}[ht]
\centering
\includegraphics[scale=0.28]{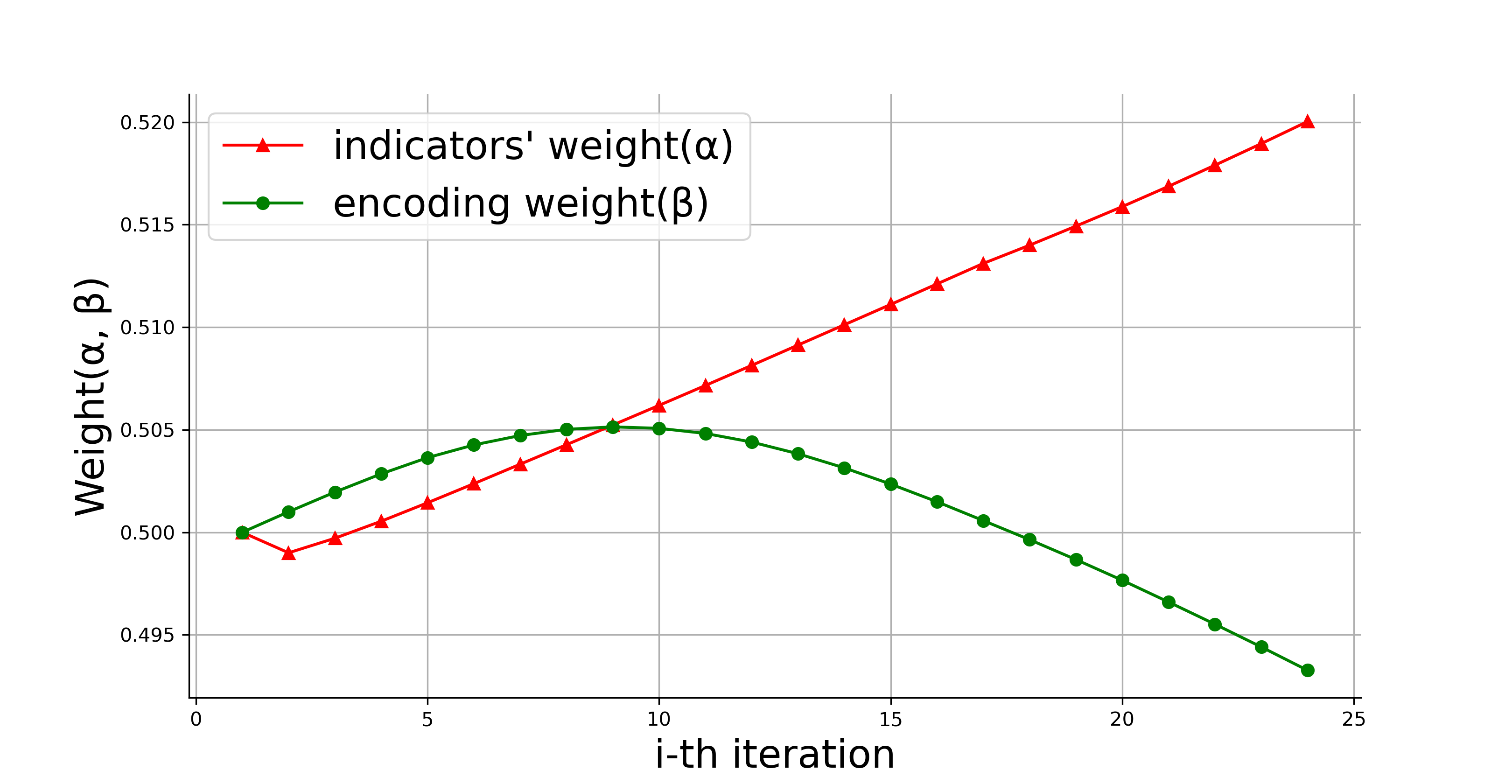}
\caption{The value of weights on each training iteration}
\label{fig_suprl}
\end{figure}
To investigate whether the three indicators selected in the supplementary sentence-level agent affect model performance, we tracked the changes of the two weights, $\alpha$ and $\beta$, during training and reported them in Fig.\ref{fig_suprl}. Both weights are initialized to 0.5 and their values change slightly during training. The weight of the three indicators does not approach 0, which demonstrates that the selected three indicators impact the model performance. Table~\ref{table_wiki} shows that the proposed model's test accuracy without these indicators is 1.3\% lower than the original proposed model. This also indicates the positive impact of these indicators on model performance. To investigate the impact of the defined transition rule, we replace the transition rule with a random selection process, in which the next state of the supplementary sentence-level agent is randomly chosen from the remaining sentences. Table~\ref{table_wiki} shows that the model accuracy based on this process is 1.7\% lower than the original model. This demonstrates that the transition rule based on variety of entities helps the proposed model alleviate the noise data effect.

\section{Conclusion}
We proposed (1) a sentence distribution estimator to alleviate the impact of noisy distant supervision labeled data for n-ary cross-sentence relation extraction; (2) 
a weaker distant supervision assumption, which considers non-consecutive sentences; and (3) a universal relation extractor, which is a hybrid model of attention mechanism and non-linear transformation layer that encodes both non-consecutive and consecutive sentence groups. The experiments showed that the proposed model reduces the impact of noisy data and achieves significantly better performance for n-ary cross sentence relation extraction compared to SotA models.
\bibliography{bibfile}
\section{Appendix}
\subsection{Algorithm}
The training procedure of the proposed model is shown in Algorithm 1:\\
\SetKwInOut{Parameter}{Parameter}
\begin{algorithm}[ht]
\caption{Model Training}
\KwIn{a set of sentence groups and relation label pairs  $\mathcal{H}=\{(g_1,r_1),(g_2,r_2),\cdots,(g_n,r_n)\}$}
\KwOut{The RE and SDE trained on the input}
\Parameter{the number of training times for RE, SDE and the whole model, $M$, $J$ and $K$, respectively}
Initialize parameters of the RE and SDE model\;
\For{$m=1\rightarrow M$} 
{
 RE receives $\mathcal{G}=\{g_1,g_2,\cdots,g_n\}$ as input\\
 RE outputs the classification results $\hat{\mathcal{R}}=\{\hat{r}_1,\hat{r}_1,\cdots,\hat{r}_n\}$\\
 calculate cross entropy loss based on the $\hat{\mathcal{R}}$ and $\mathcal{R}=\{r_1,r_2,\cdots,r_n\}$\\
 update parameters of RE model
}
\For{$k=1\rightarrow K$}
{
\For{$j=1\rightarrow J$}
{
SDE samples instances $\mathcal{G}'=\{g'_1,g'_2,\cdots,g'_i\}$ from $\mathcal{G}$ via Eq.1$\sim$3\\
do step 18$\sim$19\\
calculate reward based on the classification accuracy from RE via Eq.4\\
calculate policy gradient via Eq.5\\
update parameters of SDE model via Eq.6
}
SDE samples instances $\mathcal{G}'$ from $\mathcal{G}$\\
\For{$m=1\rightarrow M$}
{
RE receives the sampled instances $\mathcal{G}'$ as input\\
RE outputs the classification results $\hat{\mathcal{R}}$\\
calculate cross entropy loss based on the $\hat{\mathcal{R}}$ and $\mathcal{R}$\\
update parameters of RE model
}
}
return RE, SDE
\end{algorithm}
\subsection{Theoretical Analysis}
We theoretically show that the proposed model can classify correctly labeled data and remove the incorrectly labeled data.

To formalize this statement, we first formally define the training data distribution. Suppose we have true distribution $p_X(x)$ and noisy distribution $\xi$. In general, $\xi$ has zero mean. In addition, each sampled data of $\xi$ is not correlated. Then $x_p\sim p_X(x)$ is the correctly labeled data and $x_n\sim p_X(x)+\xi$ is the incorrectly labeled data. The main reason is that incorrectly labeled sentences have the same entity set with the correctly labeled sentences but the semantic information of incorrectly labeled sentences are shifted by some noise.

Then the statement is equivalent to corollary 1 below.\\
\textbf{Corollary 1.} After iteratively training SDE and RE, we have:
\begin{equation}
    P_{\theta,\gamma}(1|x_p) \gg P_{\theta,\gamma}(1|x_n)
\end{equation} where $P_{\theta,\gamma}(1|x_p)$ indicates the probability of selecting sampled $x_p$ as the positive labeled data.
To prove Corollary 1, we define two lemmas and give the proof sketch of these two lemmas\footnote{The full proof is a work in progress}:\\
\textbf{Lemma 1.} 
Let $r$ be the average reward, for RL, we have
\begin{equation}
\begin{aligned}
    \max R &\equiv \max R_{\overline{p}} + \min R_{\overline{n}}\\
    where \; &R_{\overline{p}}=R-r>0 \; R_{\overline{n}}=R-r<0\\
\end{aligned}
\end{equation}
\textbf{Proof of Lemma 1.}
The objective of RL is to maximize the expected reward, as stated in Eq.\ref{eq-r}
\begin{equation}
\begin{aligned}
R&=\sum_{\tau}R(\tau)\pi_{\theta, \gamma}(\tau)\\
&= \sum_{\tau}\left(R(\tau)-r\right)\pi_{\theta, \gamma}(\tau)\\
\end{aligned}
\label{eq-r}
\end{equation} Subtracting $r$ from $R$ is equivalent to the original reward function, because this is an unbiased estimation of expectation. Based on this, maximizing $R$ is equivalent as follows:
\begin{equation}
\begin{aligned}
    &\max R \\
    &\propto \max\left(\sum_{\tau \in \tau_p}\left(R_p(\tau)\right)\pi_{\theta, \gamma}(\tau)+\sum_{\tau \in \tau_n}\left(R_n(\tau)\right)\pi_{\theta, \gamma}(\tau)\right)\\
    &=\max\left(\sum_{\tau \in \tau_p}\left(R_p(\tau)\right)\pi_{\theta, \gamma}(\tau)\right)+\min\left(\sum_{\tau \in \tau_n}\left(R_n(\tau)\right)\pi_{\theta, \gamma}(\tau)\right)\\
    &=\max R_{\overline{p}} + \min R_{\overline{n}}\\
&where \; \tau_p\in\{\tau|R(\tau)-r>0\} \; \tau_n\in\{\tau|R(\tau)-r<0\}
\end{aligned}
\end{equation} 
$\qed$
By maximizing the first term $R_{\overline{p}}$, Lemma 1 indicates that after training the RL, the data that has higher reward ($R-r>0$) will be assigned higher probability to be selected as truly labeled data. Similarly, the data that has lower reward ($R-r<0$) will be assigned lower probability by minimizing the second term $R_{\overline{n}}$.

Since the relation extraction task is a classification problem, we use cross-entropy as the loss function of the RE model\footnote{We use binary classification for simplification, the case is the same in multi-class classification}:
\begin{equation}
\begin{aligned}
\mathcal{L} &= \mathbb{E}_{x_p\sim p_X(x)}[ylog(\hat{y}(x))+(1-y)log(1-\hat{y}(x))] \\
&+\mathbb{E}_{x_n\sim p_X(x)+\xi}[ylog(\hat{y}(x)+(1-y)log(1-\hat{y}(x))]\\
\end{aligned}
\label{loss}
\end{equation}
\textbf{Lemma 2.} After training RE model, we have
\begin{equation}
\begin{aligned}
&\min \mathcal{L} \equiv \min \mathcal{L}_r + \upsilon^2\mathcal{L}_n\\
\text{where } \; \mathcal{L}_r &= -\sum_{x\in \{x_p, x_n-\xi\}}ylog(\hat{y}(x))\\
&+(1-\hat{y}(x))log(1-y_p)p(y|x)p(x)\\
& \int \xi_i\xi_jp(\xi)d\xi = \upsilon^2\delta_{ij}
  \end{aligned}
 \label{lemma2}
\end{equation} 
$\mathcal{L}_n$ is positive definite and can be deemed as a regularization term when $\upsilon^2$ is small.

Lemma 2 indicates that the model will converge to the distribution of $x_p$ with regularization. In other words, the cross-entropy loss of $x_p$ is much lower than that of $x_n$.\\
\textbf{Proof of Lemma 2.} We expand the loss function as a Taylor series in powers of $\xi$ and substitute the Taylor expansion into the loss function. Then the loss function can be re-written as:
\begin{equation}
\begin{aligned}
\mathcal{L} &= \mathcal{L}_r + \upsilon^2 \mathcal{L}_e\\
\mathcal{L}_e &= \frac{1}{2}\sum_{x\in x_n-\xi}\left \{\left[\frac{\hat{y}-y}{\hat{y}(1-y)}\right]\frac{\partial^2 \hat{y}}{\partial x^2}\right .\\
&+\left .\left[\frac{1}{\hat{y}(1-\hat{y})}-\frac{(\hat{y}-y)(1-2\hat{y})}{\hat{y}^2(1-\hat{y})^2}\right]\left(\frac{\partial \hat{y}}{\partial x}\right)^2\right\}p(y|x)p(x)\\
\end{aligned}
\label{loss_rewrite}
\end{equation} 
where $\upsilon$ represents the amplitude of the noise, $\hat{y}$ represents the predicted relation label of $x$ by the RE model and $y$ represents the real label. 
In general, $\hat{y}$ represents the probability of predicting the correct label for $x$ should be labeled as the correct relation. As proved in previous literature, the second and third term $\mathcal{L}_e$ vanish after training the model~\cite{bishop1995training}. Then $\mathcal{L}_e$ is equivalent to:
\begin{equation}
\mathcal{L}_e = \frac{1}{2}\sum_{x\in x_n-\xi}\frac{1}{\hat{y}(1-\hat{y})}(\frac{\partial \hat{y}}{\partial x})^2p(x)
\end{equation}
Now $\mathcal{L}_e$ only has first derivatives and is positive definite. In other words, $\mathcal{L}_e = \mathcal{L}_n$ and it can be deemed as a regularizer.
$\qed$
From Lemma 1 and Lemma 2: upon completion of the iterative training of the SDE and RE models, the probability of selecting the correctly labeled data $x_p$ as positively labeled data, will end up with a comparatively larger value than the incorrectly labeled data $x_n$. That is, corollary 1 is shown to be true.
\end{document}